# The Application of ChatGPT in Responding to Questions Related to the Boston Bowel Preparation Scale

Xiaoqiang Liu[1], Yubin Wang[1], Zicheng Huang[1], Boming Xu[1], Yilin Zeng[1], Xinqi Chen[1], Zilong Wang[3,4], Enning Yang[3,4], Xiaoxuan Lei[4], Yisen Huang[1]*, Xiaobo Liu[1,2]*

1. Department of Gastroenterology, First Hospital of Quanzhou Affiliated to Fujian Medical University, Quanzhou, Fujian, China.
2. McConnell Brain Imaging Centre, Montreal Neurological Institute, McGill University, Montreal, Quebec, Canada
3. T he Neuro - Montreal Neurological Institute (MNI), McConnell Brain Imaging Centre, Department of Biomedical Engineering, Faculty of Medicine, School of Computer Science, McGill University, Montreal, Canada.
4. Mila - Quebec Artificial Intelligence Institute, Montreal, Quebec, Canada.

* means co-corresponding author: xiaobo.liu@mail.mcgill.ca and 18159503331@163.com

Abstract

**Background**: Colonoscopy, a crucial diagnostic tool in gastroenterology, depends heavily on superior bowel preparation. ChatGPT, a large language model with emergent intelligence which also exhibits potential in medical applications. This study aims to assess the accuracy and consistency of ChatGPT in using the Boston Bowel Preparation Scale (BBPS) for colonoscopy assessment. **Methods**: We retrospectively collected 233 colonoscopy images from 2020 to 2023. These images were evaluated using the BBPS by 3 senior endoscopists and 3 novice endoscopists. Additionally, ChatGPT also assessed these images, having been divided into three groups and undergone specific Fine-tuning. Consistency was evaluated through two rounds of testing. **Results**: In the initial round, ChatGPT's accuracy varied between 48.93% and 62.66%, trailing the endoscopists' accuracy of 76.68% to 77.83%. Kappa values for ChatGPT was between 0.52 and 0.53, compared to 0.75 to 0.87 for the endoscopists. **Conclusion**: While ChatGPT shows promise in bowel preparation scoring, it currently does not match the accuracy and consistency of experienced endoscopists. Future research should focus on in-depth Fine-tuning.

**Introduction**

In the realm of gastroenterology, colonoscopy is recognized as an essential diagnostic tool for colonic pathologies, heavily reliant on comprehensive bowel preparation. The importance of high-quality bowel preparation extends beyond enhancing the detection rate of colorectal adenomas and the success rate of cecal intubation; it also significantly reduces the risk of missing adenomas, with rates reported between 35%-42% in cases of inadequate preparation. [1,2]   The Boston Bowel Preparation Scale (BBPS) is widely used in clinical settings for its detailed segmental assessment. However, it is hard for

physician to understand and evaluate bowel cleanliness because of a lack of uniform assessment standards. [3]

Confronted with this challenge, the incorporation of Artificial Intelligence (AI) technology presents an innovative approach to bowel preparation scoring.[4] Particularly noteworthy are the significant achievements of Convolutional Neural Networks (CNN) in medical image analysis, exemplified by the AI system ENDOANGEL, which has brought a higher level of accuracy and consistency to bowel preparation scoring. [2] These advancements suggest that deep learning and AI could assist clinicians in improving the accuracy and consistency of bowel preparation scoring.

In the past few years, CNNs have advanced in bowel preparation assessment. However, they fall short in clinical usability and human-centered communication. In other words, most deep learning models have this hard boundary in a discriminative way instead of parametrically assigning probabilities to work together with clinicians. Conversely, ChatGPT, a large language processing model, has shown proficiency in processing medical texts and enhancing doctor-patient communication, exemplified by its ability to provide contextually informed suggestions on colonoscopy follow-up timings.[5] Furthermore, ChatGPT 4.0 has shown potential in image recognition, enhancing its multimodal analysis capabilities. However, the effectiveness of ChatGPT's clinical application remains to be tested, and advancing its clinical integration is still a challenge.

To filling the gap of AI's application to Colonscopy's field of helping clinicians, this study focuses on examining ChatGPT's utility in addressing the BBPS questions. We want to uncover ChatGPT's operational ease and ability to humanize communication in clinical environments, as well as to evaluate the significance of its image processing function in bowel preparation scoring. Our research indicates that ChatGPT has shown potential in bowel preparation scoring, and future studies should focus on further Fine-tuning.

**Materials and Methods**

**Colonoscopy Image Collection, Screening, and Categorization**

This study involved a retrospective collection of 252 colonoscopy images spanning from 2020 to 2023. For a balanced evaluation, images were evenly selected across four distinct BBPS levels (BBPS-0, 1, 2, 3), with 63 images representing each level. These images originated from two different equipment suppliers: Japan's PENTAX Corporation and Olympus Corporation.

Three seasoned endoscopy experts, each with over a decade of experience in colonoscopy, were recruited to score and categorize the images. Prior to data categorization, each expert underwent basic training and watched instructional videos about BBPS. To reduce bias from individual experts, we selected only those images that received the same score from at least two experts for testing and learning.

During the screening process, we excluded 4 images of poor quality and 3 images with completely inconsistent scoring among experts, ultimately selecting 233 images for testing and 12 images for ChatGPT's Fine-tuning ( Figure 1).

**Fine-tuning of ChatGPT**

This study investigated three distinct Fine-tuning methods aimed at augmenting ChatGPT 4.0's proficiency in recognizing colonoscopy images and adhering to relevant scoring standards. Specifically, we designed the following three models: 1. ChatGPT-T: This model used only text materials about BBPS for fine-tuning. 2. ChatGPT-1Pic: This model combined BBPS text materials with one representative image for each BBPS level, aiming to enhance the model's ability to recognize and analyze individual colonoscopy images. 3. ChatGPT-3Pic: In this version, we used BBPS text materials supplemented with three images for each BBPS level for Fine-tuning, to further strengthen the model's capability in comprehensive processing of multiple colonoscopy images. Through these customized Fine-tuning methods, we aimed to significantly improve ChatGPT's accuracy and efficiency in handling colonoscopy-related images and their scoring criteria.

**BBPS Scoring by ChatGPT, Novices, and Seniors**

The endoscopists participating in this test were divided into two groups: three novices (with less than 5 years of experience in colonoscopy) and three seniors (with 5 to 10 years of experience). During the test, these physicians and ChatGPT-T, ChatGPT-1Pic, and ChatGPT-3Pic were all required to score each image using BBPS, and each physician's response time was limited to 30 minutes to meet clinical practicality.

**Stability Assessment of ChatGPT, Novice, and Senior Scoring**

Additionally, the stability of scoring by ChatGPT, novices, and seniors was evaluated. One week later, we shuffled the order of the images and conducted a second round of tests and consistency analysis.

**Comparison with Machine Learning Models**

We compared Support Vector Machine (SVM) and Decision Tree classifiers, all based on MATLAB 2023. SVM used the fitsvm function, and Decision Tree used the fitree function, with all parameters set to default. Specifically, we used the same test and training set as ChatGPT, with 3 images per score, totaling 12 images for the training set and 233 images for the test set.

**Performance Prediction Evaluation**

When quantitative data meet normal distribution or approximate normal distribution, it is represented by $\bar{x} \pm s$. Count data is represented by frequency (%), and comparison between groups is performed using the chi-square test. To gauge accuracy, the F1 Score and Precision metrics were employed to assess the scoring accuracy of ChatGPT, seniors, and novices regarding BBPS.[6] Additionally, to evaluate their stability and consistency in two rounds of testing, we applied Cohen's kappa coefficient (with kappa values ranging from 0-0.2 considered slight, 0.21-0.4 fair, 0.41-0.6 moderate, 0.61-0.8 substantial, and 0.81-1.0 almost perfect).[7] When comparing machine learning accuracy, we similarly used Precision as the evaluation standard. All analyses were conducted using MATLAB 2023 and IBM SPSS version 23.

**Results**

**Assessing the Impact of Different Fine-tuning Methods on ChatGPT's BBPS Scoring Accuracy and F1 Score**

Regarding BBPS scoring, the initial answer accuracy of ChatGPT-T, ChatGPT-1Pic, and ChatGPT-3Pic was 48.93%, 49.79%, and 62.66%. The differences among the three groups are statistically significant ($\chi^2$=11.097, $P$=0.004, Table1). Upon pairwise comparison, it was found that the accuracy difference between ChatGPT-3Pic and both ChatGPT-T and ChatGPT-1Pic is statistically significant ($P$<0.05), while there is no significant difference in accuracy between ChatGPT-T and ChatGPT-1Pic ($P$>0.05). F1 Scores for BBPS categories 0-3 for ChatGPT-T, ChatGPT-1Pic, and ChatGPT-3Pic was computed as follows: ChatGPT-T scored 0.44, 0.34, 0.48, 0.67; ChatGPT-1Pic scored 0.46, 0.27, 0.48, 0.74; ChatGPT-3Pic scored 0.46, 0.64, 0.57, 0.75 (Figure 2). In the confusion matrices of ChatGPT-T and ChatGPT-1Pic, the highest error rate was observed for BBPS-1. In ChatGPT-3Pic's confusion matrix, the highest error rate was for BBPS-0.

**Accuracy and F1 Score of Seniors and Novices for BBPS Scoring Compared to ChatGPT-3Pic**

When compared to ChatGPT-3Pic, endoscopists demonstrated superior performance, with seniors and novices achieving an accuracy of 77.83% and 76.68% in the first round of answers. The differences among the three groups are statistically significant ($\chi^2$=16.591,$P$<0.001,Table2). Upon pairwise comparison, it was found that the accuracy difference between ChatGPT-3Pic and both seniors and novices is statistically significant ($P$<0.05), while there is no significant difference in accuracy between seniors and novices ($P>0.05$) .The F1 Scores for BBPS categories 0-3 for Seniors and novices were as follows: seniors scored 0.75, 0.69, 0.76, 0.90; novices scored 0.72, 0.74, 0.75, 0.85 (Figure 3).

**Evaluating the Consistency of BBPS Scoring: ChatGPT vs. Seniors and Novices**

To further explore the stability and consistency of endoscopists compared to ChatGPT, the previously used test images were reordered, and a second round of testing was conducted. The kappa values for ChatGPT-T, ChatGPT-1Pic, and ChatGPT-3Pic was 0.53, 0.52, and 0.53, respectively, while those for seniors and novices were 0.87 and 0.75 (Figure 4).

**Evaluating the Accuracy of Machine Learning Models in BBPS Scoring**

The precision scores for SVM in scoring BBPS categories 0-3 were 0.17, 0.32, 0, 0, and for the Decision Tree, they were 0.21, 0, 0.35, 0 (Figure 5). The precision scores for SVM and Decision Tree were lower than those for ChatGPT-3Pic (0.53, 0.60, 0.61, 0.70).

Table 1 Comparison of the Accuracy of BBPS Among ChatGPT-T, ChatGPT-1Pic, and ChatGPT-3Pic Groups

|  | ChatGPT-T | ChatGPT-1Pic | ChatGPT-3Pic | $\chi^2$ | $P$ value |
|---|---|---|---|---|---|
| Right [n (%)] | 114(48.93)[a] | 116(49.7)[a] | 146(62.66)[b] | 11.097 | 0.004 |
| Wrong [n (%)] | 119(48.9)[a] | 117(50.21)[a] | 87(37.34)[b] | | |

$P$ value for $\chi^2$ test comparing the Accuracy of BBPS Among ChatGPT-T, ChatGPT-1Pic, and ChatGPT-3Pic Groups. "a" indicates groups that do not have a significant difference. "b" indicates groups that have a significant difference statistically when compared to other groups marked with "a".

Table 2 Comparison of the Accuracy of BBPS Among ChatGPT-3Pic, Seniors and Novices

|  | ChatGPT-3Pic | Seniors(mean) | Novices(mean) | $\chi^2$ | $P$ value |
|---|---|---|---|---|---|
| Right [n (%)] | 146(62.66)[b] | 181.34(77.83)[a] | 178.66(76.68)[a] | 16.591 | <0.001 |

| | | | |
|---|---|---|---|
| Wrong [n (%)] | 87(37.34)[b] | 51.66(22.17)[a] | 54.34(23.32)[a] |

*P* value for χ² test comparing the Accuracy of BBPS Among ChatGPT-3Pic, Seniors and Novices. "a" indicates groups that do not have a significant difference. "b" indicates groups that have a significant difference statistically when compared to other groups marked with "a".

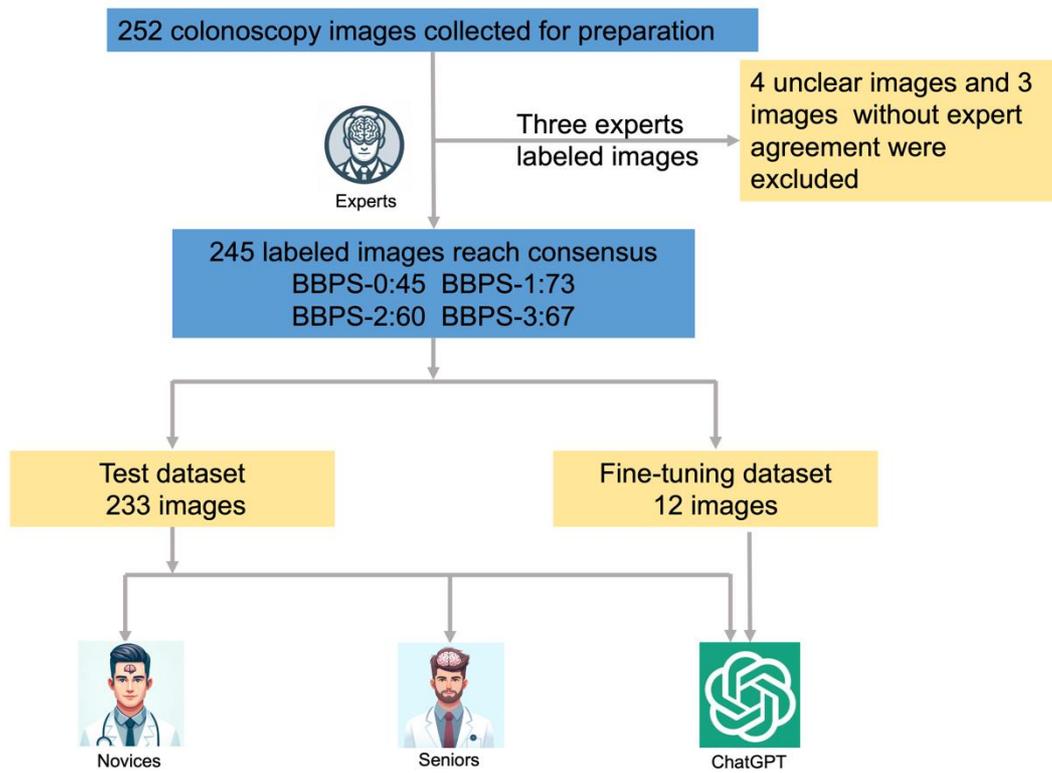

Figure 1: The flowchart provides a detailed depiction of the selection, labeling, and testing process for colonoscopy images. Initially, 252 images were screened, out of which 7 were excluded due to insufficient clarity or lack of consensus among experts. Consequently, 245 images were retained for the study. Of these, 233 images formed the test dataset, which was evaluated using the BBPS by both novices and seniors, as well as ChatGPT. The remaining 12 images comprised a Fine-tuning dataset specifically designed for ChatGPT.

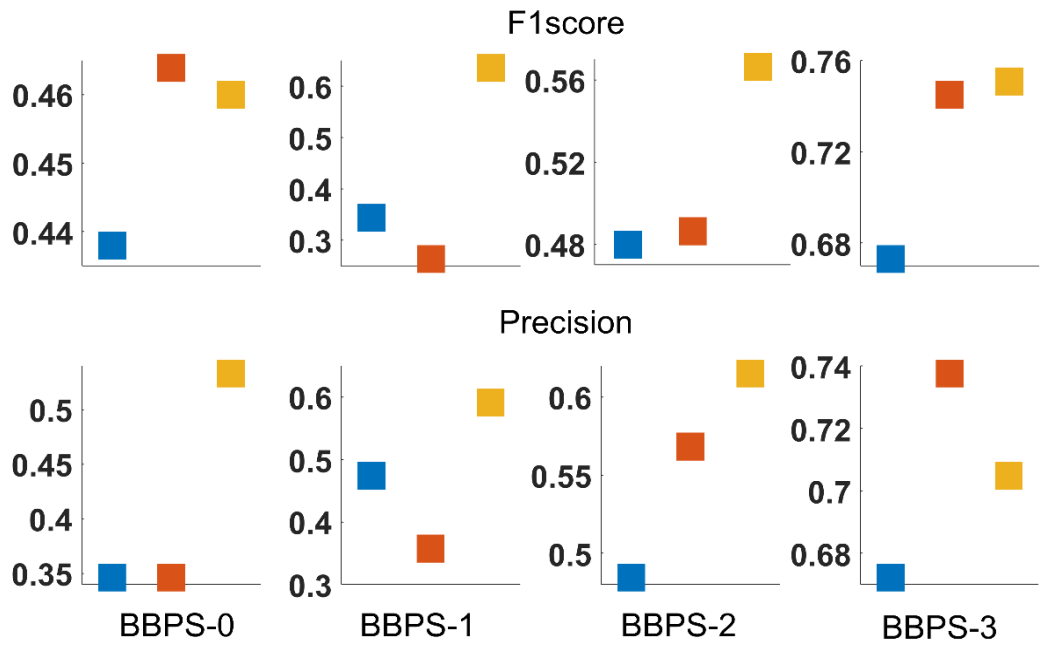

Figure 2: The Impact of Different Fine-tuning on ChatGPT's Initial Prediction of BBPS Scores. The chart uses blue to represent ChatGPT-T (BBPS text-only Fine-tuning), red to denote ChatGPT-1Pic (BBPS text combined with single-image Fine-tuning), and yellow for ChatGPT-3Pic (BBPS text complemented with multiple-image Fine-tuning).

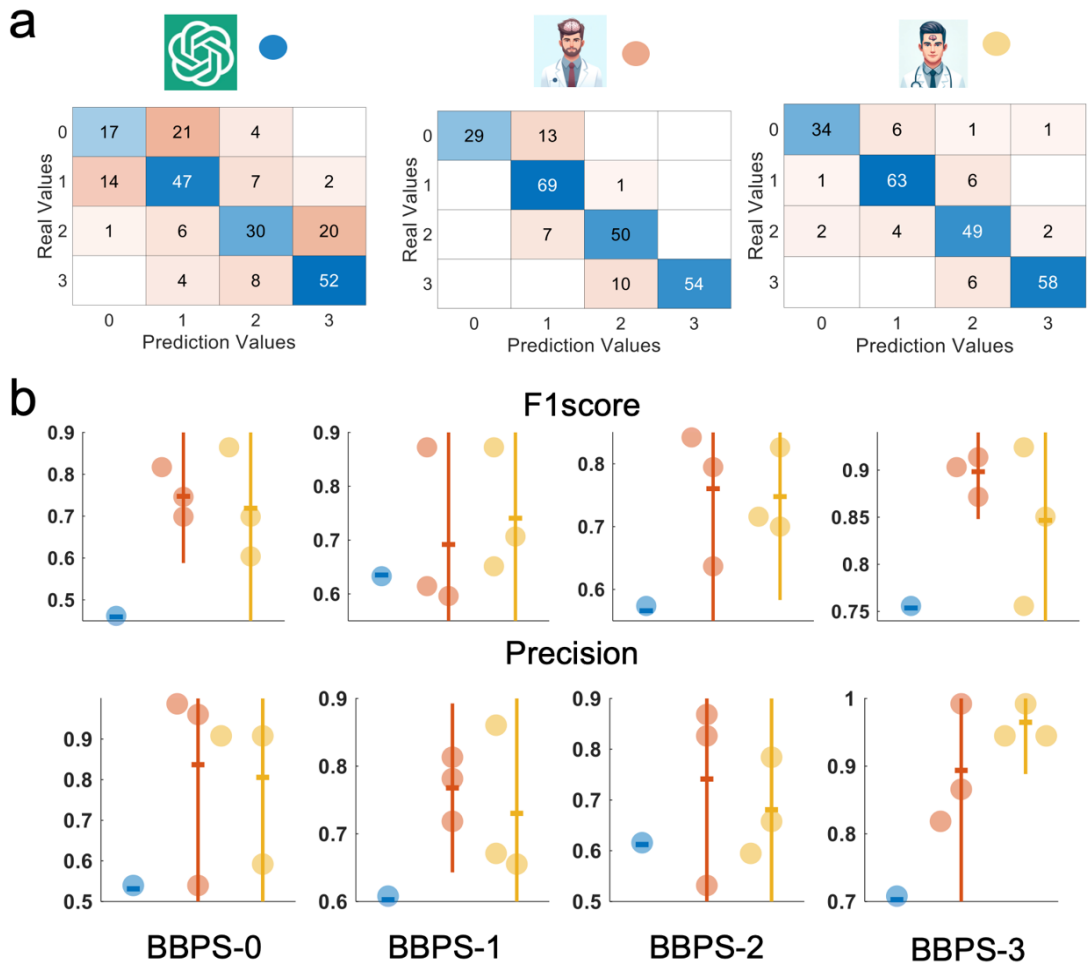

Figure 3: The Initial BBPS Scoring by ChatGPT-3Pic, Seniors, and Novices. A. The confusion matrix comparing the scores given by ChatGPT-3Pic, seniors, and novices for bowel cleanliness images against the BBPS gold standard. B. Extraction of the F1-scores and Precision for the corresponding BBPS scores of ChatGPT-3Pic, seniors, and novices from the confusion matrix. (Blue dots represent ChatGPT-3Pic, red dots denote seniors, and yellow dots indicate novices).

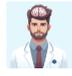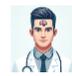
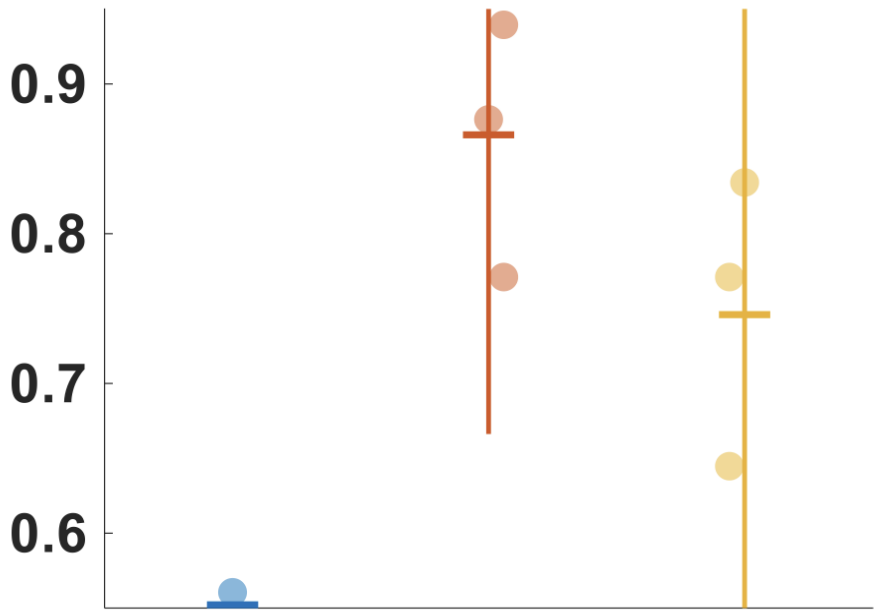

Figure 4: Stability of the Second Discrimination by ChatGPT-3Pic, Seniors, and Novices. (Blue dots represent ChatGPT-3Pic, red dots denote seniors, and yellow dots indicate novices).

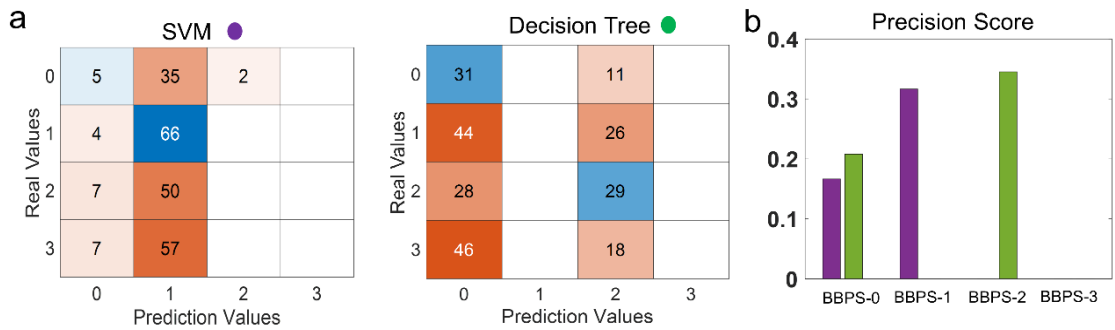

Figure 5: Accuracy Achieved Using SVM and Decision Tree Predictive Models. The predictive outcomes from both classifiers are equivalent to those of a random model.

You

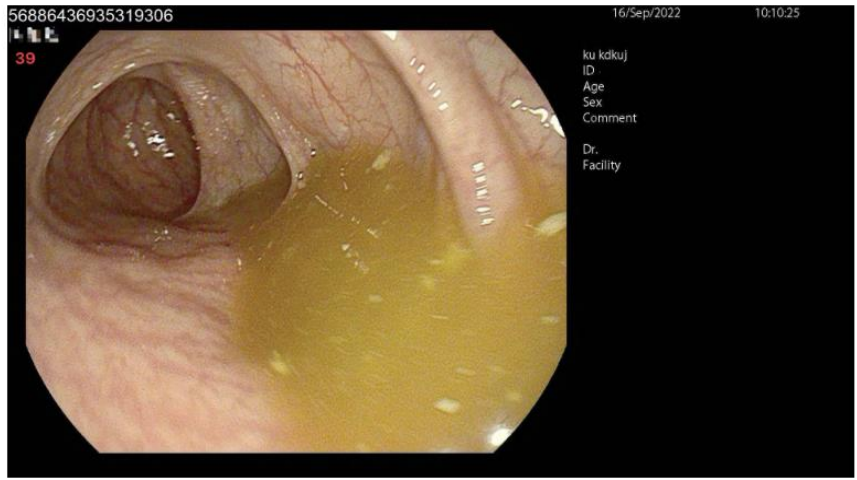

Assuming you are a gastrointestinal specialist who can identify endoscopic images of the gastrointestinal tract, the scoring criteria for the Boston Bowel Preparation Scale are as follows: 0 points for an unprepared segment of the colon with mucosa not visualized due to the presence of solid stool; 1 point for partially visible colon segments with the presence of fecal liquid and semi-solid stool; 2 points for a segment with low fecal fluid content allowing good visualization of the mucosa; 3 points for excellent visualization of the mucosa without the presence of liquid remnants. I will send you an image, and I'd like you to analyze it using the Boston Bowel Preparation Scale. Please first tell me the score, and then provide the basis for your judgmen

ChatGPT

Upon reviewing the endoscopic image provided, I would assign a score of 2 according to the Boston Bowel Preparation Scale. This score suggests that the bowel preparation is adequate, with a few minor limitations:

- The mucosa is mostly visible, which allows for a reasonable examination of the colonic lining.
- There is some fecal fluid present, but it does not significantly impede the view of the mucosa.
- The vascular pattern of the colon can be identified, which is important for the evaluation of mucosal integrity and identification of any pathological lesions.

The image does not meet the criteria for a score of 3, which requires excellent visualization of the mucosa without the presence of liquid remnants. A score of 1 would indicate that the mucosa is only partially visible due to the presence of more fecal material, which is not the case here. Lastly, a score of 0 is characterized by an unprepared colon with mucosa not visualized because of solid stool, which also does not apply to this image.

Figure 6: ChatGPT's Evaluation and Rationale for BBPS Scoring of Colonoscopy Images.

**Discussion**

This research explores the application of ChatGPT in addressing questions related to the BBPS, aiming to reveal its operational convenience and humanized communication ability in a clinical setting, as well as to assess the value of its image processing capability in bowel preparation scoring. The findings suggest that ChatGPT exhibits potential in the domain of bowel preparation scoring, and future research should focus on further fine-tuning.

**Application of ChatGPT in BBPS Scoring**

The advancement of ChatGPT suggests a pivotal moment in the Turing Test's history, casting doubt on whether AI can supplant human expertise, albeit with the assistance of human experts. [8] With this revolutionary cange, we need to rethink can we automate the jobs in the field of medical assessment. In particular, this study is the first to explore the application and efficacy of ChatGPT in BBPS scoring, highlighting its potential in this context. In the first round of testing, ChatGPT achieved an accuracy of 62.66%. Although ChatGPT offers a rapid, automated method for processing colonoscopy images, it falls short of the accuracy achieved by experienced endoscopists. The confusion matrix results indicate that ChatGPT had a higher error rate in distinguishing between BBPS-1 and BBPS-0, underscoring the challenges ChatGPT faces in interpreting complex medical images and performing detailed scoring tasks. ChatGPT typically relies on a large amount of high-quality training data to improve its performance. However, the complexity and diversity of colonoscopy images require ChatGPT to recognize varying degrees of bowel preparation and consider patient-related variables, such as differences in intestinal anatomy.

The kappa value assessment shows that ChatGPT's performance in consistency and stability is acceptable but falls short compared to experienced endoscopists. This reflects the limitations of ChatGPT in performing long-term and consistent medical image assessments, especially in complex tasks requiring detailed judgment. This suggests that while ChatGPT can serve as an auxiliary tool for preliminary analysis, the role of human experts is indispensable in clinical decision-making that requires precision and consistency.

**Humanized Human-Machine Interaction with ChatGPT**

CNN are primarily used for image processing and recognition and lack the capability to process and understand natural language, thus failing to provide a humanized human-machine communication experience. Our study found that, compared to CNNs, ChatGPT shows potential in providing a humanized interaction experience in the medical field. For instance, in BBPS scoring, ChatGPT not only provided specific scores but also detailed explanations for the basis of these scores and the reasons for not assigning other scores (Figure 6). Research indicates that ChatGPT can understand and respond to medical professionals' queries and present complex medical information in an easily comprehensible manner, suitable for non-professionals. For example, it can transform complex medical reports into formats easier for patients to understand, or generate educational health information to help patients better comprehend their health conditions and treatment options. [9] In gastroenterology, studies evaluating the quality of ChatGPT-generated answers to common questions about colonoscopy found that ChatGPT's answers were similar to non-AI answers in terms of ease of understanding, scientific sufficiency, and satisfaction, with very low textual similarity, indicating ChatGPT's originality. [10] These studies collectively suggest that ChatGPT excels in understanding complex medical texts and meeting user needs, but further research and optimization are needed to improve the accuracy of personalized interactions.

**The Importance of Endoscopists and the Role of Professional Training**

The good performance of endoscopists in accuracy and consistency underscores the importance of professional judgment and clinical experience in colonoscopy. Our study shows that even relatively inexperienced endoscopists can effectively apply BBPS for accurate bowel preparation scoring, consistent with the findings of Zhou et al., which may be attributed to systematic professional training and continuous clinical practice. [2] Endoscopists can consider subtle differences in images and specific circumstances of the clinical environment, which ChatGPT currently cannot replicate.

**Future Research Directions**

Future research should explore how to improve ChatGPT's training, which is a key area. For instance, using larger and more diverse image datasets and customized GPT models could further enhance these systems' learning and adaptability. [11] In our current study, ChatGPT-3Pic showed higher accuracy compared to ChatGPT-1Pic, likely due to more images used for Fine-tuning. This suggests that providing richer background information and data can significantly improve the performance of customized GPTs in specific tasks.

Additionally, future studies should investigate collaborative working models between ChatGPT and other GPTs and endoscopists. For example, customized GPTs could be used for preliminary screening and scoring of colonoscopy images, followed by detailed review and final confirmation by endoscopists. Such a collaborative approach could optimize the workflow of colonoscopy examinations, improve overall efficiency, and maintain a high level of accuracy and safety in the scoring process.

**Limitations of ChatGPT**

While ChatGPT offers substantial benefits in broad information searching and rapid query response, it also presents certain unavoidable limitations and specific challenges that merit further discussion and research:

1. Limitations in answering medical-related questions: In practical applications,

ChatGPT shows limitations in answering certain medical questions. [12] For example, in this study, ChatGPT-1Pic refused to answer questions four times due to sensitive medical terminology related to BBPS, requiring re-prompting to continue. This phenomenon may stem from ChatGPT's built-in guidelines, aimed at preventing the provision of information that could be misconstrued as medical advice. While this conservative response strategy helps prevent misunderstandings and misuse, it also limits ChatGPT's application in answering complex medical questions.

2. Variability in responses: ChatGPT's responses to the same question can vary. For instance, studies have noted that ChatGPT may produce different responses each time, even when given the same radiology report and prompt conditions. [13] This variability in responses may be due to the generative model's characteristics, capable of interpreting and answering the same question from multiple perspectives. However, this uncertainty can pose risks in medical applications, especially in diagnosis and treatment recommendations, where accuracy and consistency are crucial.

3. Data privacy and security issues: With the application of ChatGPT in the medical field, concerns regarding data privacy and security gain prominence. Ensuring confidentiality and security of sensitive medical information is crucial when processing it. Additionally, caution is needed against the risk of ChatGPT misleading medical diagnoses and preventing its use as definitive medical advice in inappropriate circumstances. [14]

4. Limitations in knowledge updating: Due to ChatGPT's knowledge updates being limited to April 2023, it may be limited in handling the latest medical information. [15] In the medical field, new research findings, treatment methods, and clinical guidelines are updated rapidly. Therefore, ChatGPT may be inaccurate or outdated when answering questions involving the latest medical developments or research. This is particularly important for doctors and medical researchers, who rely on the latest information to make the best clinical decisions.

5. Limitations in the number of responses: ChatGPT 4.0 can only answer 40 questions every three hours, and users must wait three hours to continue getting answers once this limit is reached. The usage limit of ChatGPT 4.0 could impact medical professionals, as in our study where we could not upload more images for Fine-tuning, affecting our final research results. Especially in emergency situations, this limitation could affect medical professionals' access to and utilization of important information at critical moments.

**Limitations of the Experimental Method**

Although this study offers initial insights into ChatGPT's application in bowel preparation scoring, it is not without its limitations. Future research should consider larger and more diverse datasets and testing in different clinical settings to further validate the effectiveness and reliability of ChatGPT in this field.

**Conclusion**

Our study results indicate that although ChatGPT shows potential in bowel preparation scoring, its current accuracy and consistency are not yet sufficient to replace professional endoscopists. More Fine-tuning are key to enhancing ChatGPT's application value in this field.